# Combinations of distributional regression algorithms with application in uncertainty estimation of corrected satellite precipitation products

Georgia Papacharalampous[1], Hristos Tyralis[2,*], Nikolaos Doulamis[3], Anastasios Doulamis[4]

[1] Department of Topography, School of Rural, Surveying and Geoinformatics Engineering, National Technical University of Athens, Iroon Polytechniou 5, 157 80 Zografou, Greece (papacharalampous.georgia@gmail.com, gpapacharalampous@hydro.ntua.gr, https://orcid.org/0000-0001-5446-954X)

[2] Department of Topography, School of Rural, Surveying and Geoinformatics Engineering, National Technical University of Athens, Iroon Polytechniou 5, 157 80 Zografou, Greece (montchrister@gmail.com, hristos@itia.ntua.gr, https://orcid.org/0000-0002-8932-4997)

[3] Department of Topography, School of Rural, Surveying and Geoinformatics Engineering, National Technical University of Athens, Iroon Polytechniou 5, 157 80 Zografou, Greece (ndoulam@cs.ntua.gr, https://orcid.org/0000-0002-4064-8990)

[4] Department of Topography, School of Rural, Surveying and Geoinformatics Engineering, National Technical University of Athens, Iroon Polytechniou 5, 157 80 Zografou, Greece (adoulam@cs.ntua.gr, https://orcid.org/0000-0002-0612-5889)

* Corresponding author

**Abstract**: To facilitate effective decision-making, precipitation datasets should include uncertainty estimates. Quantile regression with machine learning has been proposed for issuing such estimates. Distributional regression offers distinct advantages over quantile regression, including the ability to model intermittency as well as a stronger ability to extrapolate beyond the training data, which is critical for predicting extreme precipitation. Therefore, here, we introduce the concept of distributional regression in precipitation dataset creation, specifically for the spatial prediction task of correcting satellite precipitation products. Building upon this concept, we formulated new ensemble learning methods that can be valuable not only for spatial prediction but also for other prediction problems. These methods exploit conditional zero-adjusted probability distributions estimated with generalized additive models for location, scale and shape (GAMLSS), spline-based GAMLSS and distributional regression forests as well as their ensembles (stacking based on quantile regression and equal-weight averaging). To identify the most effective methods for our specific problem, we compared them to benchmarks using a large, multi-source precipitation dataset. Stacking was shown to be superior to individual methods at most quantile levels when evaluated with the quantile loss function. Moreover, while the relative ranking of the methods varied across different quantile levels, stacking methods, and to a lesser extent mean combiners, exhibited lower variance in their performance across different quantiles compared to individual methods that occasionally ranked extremely low. Overall, a task-specific combination of multiple distributional regression algorithms could yield significant benefits in terms of stability.

## 1. Introduction

Merging gauge-measured data with raw or post-processed satellite products (Baez-Villanueva et al., 2020; Papacharalampous et al. 2023a) is an important spatial prediction problem in geoinformation and earth observation engineering. Indeed, by dealing with this problem, we can create products that are both as spatially dense as needed and more accurate than pre-existing (raw or post-processed) satellite products, thereby facilitating improved input information for our technical applications.

The respective efforts for creating precipitation products have been largely benefitted from the machine learning literature (Abdollahipour et al., 2022; Hengl et al., 2018; Hu et al., 2019). The extant literature has primarily focused on point predictions. In this context, precipitation predictions at a specific location are issued as single values aiming to be as close as possible to the actual precipitation value at that point. An important topic in such spatial prediction settings is that of feature engineering for the exploitation of both spatial and temporal information (Hengl et al., 2018; Kossieris et al., 2024; Papacharalampous et al., 2023a; Sekulić et al., 2020; Villanueva et al., 2020). Again in such settings, to gain a deeper understanding of algorithm properties, researchers often evaluate multiple algorithms (Papacharalampous et al., 2023a; 2023b). Nonetheless, numerous machine learning algorithms and concepts remain unexploited in the same efforts and in the efforts for dealing with similar spatial prediction problems. This holds despite the useful advances that these algorithms and concepts could bring, especially in relatively new and unexplored endeavours, such as the estimation of the predictive uncertainty and the focus on intermittency and extreme events (Papacharalampous and Tyralis, 2022; Tyralis and Papacharalampous, 2024).

Predictive uncertainty estimation can be defined as the process of issuing predictions in the form of probability distributions (as opposed to the point predictions, which usually refer to the mean of a predictive distribution) and is needed for effective decision making (Gneiting & Raftery, 2007). Under this well-established view, in precipitation dataset creation via data merging, a predictive probability distribution should be issued for every point of interest in space. Moreover, it is important that probability distributions model characteristic properties of precipitation such as intermittency (precipitation is a non-negative variable with mass at zero) and extremes. At the same time, probabilistic predictions should be good in absolute terms (Fissler & Ziegel, 2019).

Given the above, our aim was to advance the topic of estimating predictive uncertainty of precipitation predictions with a focus on: (a) spatial prediction settings, particularly those involving the integration of remote sensing data; and (b) distinctive properties of precipitation. To achieve successful algorithmic formulations, we built upon distributional regression, a machine learning concept unexploited currently in (precipitation) dataset creation through data merging. While typical regression (Efron & Hastie, 2016; Hastie et al., 2009; James et al., 2013) models how the mean of the response variable changes with the changes of the predictor variables, distributional regression finds how the parameters of a probability distribution describing the response variable change with the changes of the predictor variables (Kneib et al., 2023; Rigby & Stasinopoulos, 2005; Tyralis and Papacharalampous, 2024). We used three types of distributional regression models, i.e. generalized additive models for location, scale and shape (GAMLSS) with linear predictors (Rigby & Stasinopoulos, 2005), spline-based

GAMLSS (Eilers & Marx, 1996; Eilers et al., 2015; Stasinopoulos et al., 2023) and distributional regression forests (Schlosser et al., 2019). Precipitation was modelled by two probability distributions, namely the zero adjusted Inverse Gaussian distribution (Heller et al., 2006) and zero adjusted Gamma distribution (Stasinopoulos et al., 2023) that can model intermittent random variables.

Two key advancements are presented in this work:

a. With respect to the existing methods for the engineering task of estimating predictive uncertainty when merging gauge-based and remote sensing precipitation datasets (e.g., those by Bhuiyan et al., 2018; Glawion et al., 2023; Papacharalampous et al., 2024a, 2024b; Tyralis et al., 2023; Zhang et al., 2022) which are all quantile regression or generative model-based and, thus, non-parametric, we propose new methods that can model intermittency better as well as extrapolate at high quantiles.

b. From a machine learning methodological point of view, we propose ensemble learning of distributional regression algorithms for zero-adjusted probability distributions based on quantile-loss optimization, that outperforms individual algorithms with respect to quantile scoring functions (Gneiting, 2011) and quantile scoring rules (Gneiting & Raftery, 2007). The concept of ensemble learning (see the reviews by Papacharalampous & Tyralis, 2022; Sagi & Rokach, 2018; Tyralis & Papacharalampous, 2024; Wang et al., 2023) was exploited, not only by simply averaging distributional regression algorithms (Lichtendahl et al., 2013) but also by using linear quantile regression algorithms as combiners (van der Laan et al., 2007; Wolpert, 1992; Yao et al., 2018), to improve predictive performance with respect to individual implementations of distributional regression algorithms.

The methods were applied to a remote sensing data-merging problem. The dataset included gauge-based measurements and satellite data from the Precipitation Estimation from Remotely Sensed Information using Artificial Neural Networks (PERSIANN) and the GPM Integrated Multi-satellitE Retrievals (IMERG) datasets. All these data are of monthly temporal resolution, they span the 2000-2015 period and cover spatially the contiguous US (CONUS).

The remaining parts of the paper are organized as follows. Section 2 describes the new methods for predictive uncertainty quantification and their various components. Section 3 summarizes information about the data on which the methods were compared in merging precipitation datasets, as well as the application protocol and how the latter facilitates investigations in terms of extremes. Section 4 presents the results. Section 5 provides in depth discussions on the methods and their usefulness, and proposes open topics for future research. Section 6 provides the conclusions.

## 2. Methods and algorithms

### 2.1 Distributional regression

Let $\underline{x}$ be a vector of predictor variables and $\underline{y}$ be the dependent variable, where we underline variables to denote that they are random. Let also the cumulative distribution function (CDF) of the random variable $\underline{y}$ conditional on the realization $x$ be $F_{\underline{y}|\underline{x}}(\boldsymbol{\theta}_0)$ where $\boldsymbol{\theta}_0$ is the parameter vector of the CDF. In the general formulation of distributional regression, a parametric model $m(\underline{x}, \boldsymbol{a}_0)$ that predicts the parameter $\boldsymbol{\theta}_0$ of $F_{\underline{y}|\underline{x}}$ has to be estimated. In practice, to estimate $\boldsymbol{a}_0$, one has the realizations $x_i$ and $y_i$, $i = 1, \ldots, n$ of the random variables $\underline{x}$ and $\underline{y}$. The estimator of $\boldsymbol{a}_0$ is (Gneiting & Raftery, 2007)

$$\hat{\boldsymbol{a}}_n = \arg \min_{\boldsymbol{a} \in A}(1/n) \sum_{i=1}^{n} L(m(x_i, \boldsymbol{a}), y_i) \qquad (1)$$

where $L(z, y)$ is a suitable loss function for the task at hand.

Then, the trained algorithm can predict the parameter $\boldsymbol{\theta}_0$, that depends on $\underline{x}$ when new values of the predictor variable are given and, thus, it can also predict the probability distribution of the random variable $\underline{y}$ conditional on $x$.

*2.1.1 Linear Generalized Additive Models for Location, Scale and Shape (GAMLSS)*

GAMLSS is the first model of this category (Rigby and Stasinopoulos 2005) that allowed for a parameter vector $\boldsymbol{\theta}_0$ with three or more elements. The loss function used in GAMLSS is a penalized likelihood function, while the parameters are linear functions of the covariates and modelled separately one from another. The R software implementation of GAMLSS by Stasinopoulos and Rigby (2024) was used. Herein, GAMLSS was applied with two probability distributions (see Section 2.1.4). Its remaining parameters were set to their defaults in Stasinopoulos and Rigby (2024).

*2.1.2 GAMLSS with P-splines*

To allow for more flexibility in GAMLSS, the parameters of the conditional CDF could be modelled by *P*-splines (Stasinopoulos et al., 2023). *P*-splines constitute a flexible framework that builds on *B*-splines. *B*-splines traditionally are made from connected polynomial pieces. These pieces join at specific points called knots, which traditionally are spaced evenly. However, this approach limits how smooth and flexible the final curve can be. *P*-splines address this by using a large number of knots and adding a difference penalty on coefficients of adjacent *B*-splines (Eilers and Marx, 1996; Eilers et al., 2015). The R software implementation of GAMLSS with *P*-splines by Stasinopoulos and Rigby (2024) was used. Herein, splines were applied with the lambda parameter equal to 1 000. Their remaining parameters were set to their defaults in Stasinopoulos and Rigby (2024).

*2.1.3 Distributional regression forests*

Distributional regression forests (Schlosser et al., 2019) are a variant of random forests (Breiman, 2001) that allows modelling the parameters of the conditional CDF using an ensemble of decision trees. Random forests are the average of the ensemble of decision trees that were trained previously with resampled data (Breiman, 2001). The ensemble of decision trees allows for predictions less variable compared to predictions of individual decision trees. Distributional regression forests are based on the random forests idea of using an ensemble of decision trees. The difference is that decision trees are trained by splitting the data to more homogeneous groups with respect to a probability distribution that allows modelling the CDF's parameters (Schlosser et al., 2019). The R software implementation of distributional regression trees by Schlosser et al. (2021) was used. Herein, distributional regression forests were applied with 100 trees. Their remaining parameters were set to their defaults in Schlosser et al. (2021).

*2.1.4 Probability distributions*

We modelled the dependent variable using two probability distributions, namely the zero adjusted Inverse Gaussian (ZAIG) distribution (Heller et al. 2006) and the zero adjusted Gamma (ZAGA) distribution (Stasinopoulos et al. 2023). The density function of the ZAIG distribution is (Heller et al., 2006)

$$f_{\text{ZAIG}}(y|\mu,\sigma,\nu) = \begin{cases} \nu, y = 0 \\ (1-\nu)(1/\sqrt{2\pi\sigma^2 y^3})\exp(-(y-\mu)^2/(2\mu^2\sigma^2 y)), y > 0 \end{cases} \quad (2)$$

and the density of the ZAGA distribution is (Stasinopoulos et al., 2023)

$$f_{\text{ZAGA}}(y|\mu,\sigma,\nu) = \begin{cases} \nu, y = 0 \\ (1-\nu)(\frac{1}{(\sigma^2\mu)^{1/\sigma^2}}\frac{y^{(1/\sigma^2)-1}\exp(-y/(\sigma^2\mu))}{\Gamma(1/\sigma^2)}), y > 0 \end{cases} \quad (3)$$

Both probability distributions have parameters $\mu > 0$, $\sigma > 0$ and $0 < \nu < 1$ and are members of the exponential family of distributions. They are mixed distributions that allow a mass at zero and are continuous when $y > 0$.

## 2.2   Overview of distributional regression algorithms

The following six individual distributional regression algorithms were employed:

o   GAMLSS with the ZAIG conditional distribution (GAMLSS-ZAIG).
o   GAMLSS with the ZAGA conditional distribution (GAMLSS-ZAGA).
o   Spline-based GAMLSS with the ZAIG conditional distribution (GAMLSS-ZAIG-Splines).
o   Spline-based GAMLSS with the ZAGA conditional distribution (GAMLSS-ZAGA-Splines).
o   Distributional regression forests with the ZAIG conditional distribution (DRF-ZAIG).
o   Distributional regression forests with the ZAGA conditional distribution (DRF-ZAGA).

Among the above, the latter four were used as base learners in the formulation of 11 ensemble learners. The six simple ensemble learners (described in detail in Section 2.3) are the following:

o   Mean combiner of GAMLSS-ZAIG-Splines and GAMLSS-ZAGA-Splines.
o   Mean combiner of DRF-ZAIG and DRF-ZAGA.
o   Mean combiner of GAMLSS-ZAIG-Splines and DRF-ZAIG.
o   Mean combiner of GAMLSS-ZAGA-Splines and DRF-ZAGA.
o   Mean combiner of GAMLSS-ZAIG-Splines, GAMLSS-ZAGA-Splines, DRF-ZAIG and DRF-ZAGA.
o   Median combiner of GAMLSS-ZAIG-Splines, GAMLSS-ZAGA-Splines, DRF-ZAIG and DRF-ZAGA.

Five stacked generalization ensemble learners (described in detail in Section 2.3) were also formulated. These are the following:

o   Stacked generalization of GAMLSS-ZAIG-Splines and GAMLSS-ZAGA-Splines.
o   Stacked generalization of DRF-ZAIG and DRF-ZAGA.
o   Stacked generalization of GAMLSS-ZAIG-Splines and DRF-ZAIG.
o   Stacked generalization of GAMLSS-ZAGA-Splines and DRF-ZAGA.
o   Stacked generalization of GAMLSS-ZAIG-Splines, GAMLSS-ZAGA-Splines, DRF-ZAIG and DRF-ZAGA.

In total, 17 algorithms (six individual, six based on simple averaging and five based on stacking) were compared.

## 2.3 Ensemble learners

We formulated stacked generalization ensemble learners that combine the independent probabilistic predictions of distributional regression algorithms. For this formulation, the employment of the linear quantile regression algorithm (Koenker, 2005; Koenker & Bassett, 1978) as the combiner is proposed for the first time. Under this approach, the quantile regression algorithm delivers the $\tau$-quantile prediction of the ensemble learner by using as predictor variables the $\tau$-quantile predictions extracted from the distributional predictions of the base learners. The linear quantile regression algorithm is a linear model trained to minimize the average quantile loss through the quantile loss (scoring) function (Gneiting, 2011; Koenker & Bassett, 1978; Saerens, 2000; Thomson, 1979). This function is defined, as in (Koenker & Bassett, 1978)

$$L_\tau(z, y) \coloneqq (z - y)(\mathbb{I}(z \geq y) - \tau), \qquad (4)$$

where $\tau$, $y$ and $z$ are the quantile level, the observation and the prediction, respectively, and $\mathbb{I}(A)$ is the indicator function. The latter is equal to 1 when the event $A$ realizes and equal to 0 otherwise. The quantile loss function is also used to evaluate predictions of $\tau$-quantiles (Gneiting, 2011).

More precisely, the five ensemble learners can be implemented using the following pseudo algorithm (see also Figure 1), which assumes that $n$ samples are available for the training:

- Step 1: Randomly split the $n$ samples into sets 1 and 2 containing $n_1$ and $n_2$ samples, respectively, with $n_1 + n_2 = n$.
- Step 2: Train $k$ distributional regression algorithms on set 1 to obtain predictive distributions for set 2. Let the predictions in set 2 be notated with $f_1, \ldots, f_k$.
- Step 3: Extract the predictive quantile at any level of interest $\tau$ from each predictive distribution $f_1, \ldots, f_k$. Let the predictive quantiles be denoted with $q_{1,\tau}, \ldots, q_{k,\tau}$.
- Step 4: Train the combiner to minimize the average quantile score at level $\tau$ using $q_{1,\tau}, \ldots, q_{k,\tau}$ on set 2 as predictor variables.

- Step 5: Retrain the $k$ distributional regression algorithms $1, \ldots, k$ on the union of set 1 and set 2 to issue predictive distributions for new samples, which compose the test set. Let these predictive distributions be notated with $f_{\text{upd},1}, \ldots, f_{\text{upd},k}$.
- Step 6: Extract the predictive quantile at the level $\tau$ from each predictive distribution $f_{\text{upd},1}, \ldots, f_{\text{upd},k}$. Let these predictive quantiles be denoted with $q_{\text{upd},1,\tau}, \ldots, q_{\text{upd},k,\tau}$.
- Step 7: Produce quantile predictions for the test samples by applying the trained combiner of step 4 with $q_{\text{upd},1,\tau}, \ldots, q_{\text{upd},k,\tau}$ as the predictor variables.

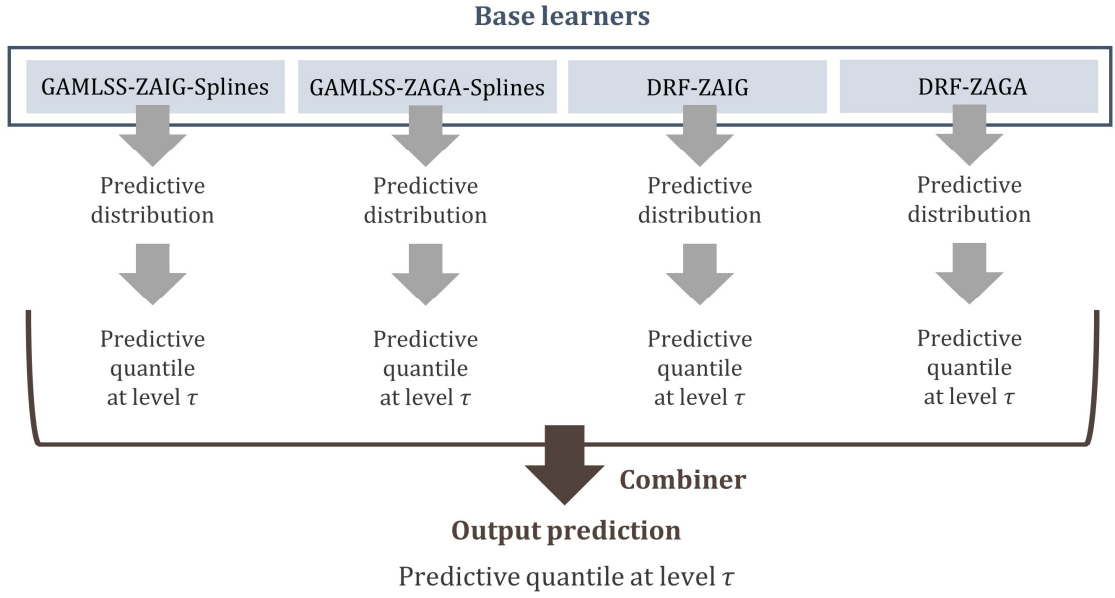

Figure 1. Formulation of the ensemble learning algorithms using GAMLSS-ZAIG-Splines, GAMLSS-ZAGA-Splines, DRF-ZAIG and DRF-ZAGA as their base learners. The remaining ensemble learning algorithms were formulated in a similar way.

To provide benchmarks for the stacked generalization ensemble learners, additionally to the individual algorithms listed in Section 2.2, six simple ensemble learners were also formulated (see also Section 2.2). Each simple ensemble learner simply computes the mean or the median of the $\tau$-quantile predictions extracted from the distributional predictions of the base learners.

## 3. Datasets and application

### 3.1 Datasets

We applied the algorithms (see Section 2.2) for estimating predictive uncertainty in the problem of merging precipitation data from multiple databases. In the application, we used gauge-measured precipitation data for the 1 421 locations shown in Figure 2 and the years 2001–2015, remote sensed precipitation data for the 0.25° × 0.25° grids shown in Figure 3 and the same years, and elevation data for the 1 421 locations shown in Figure 2 (because this latter variable usually consists an informative predictor of hydrometeorological variables).

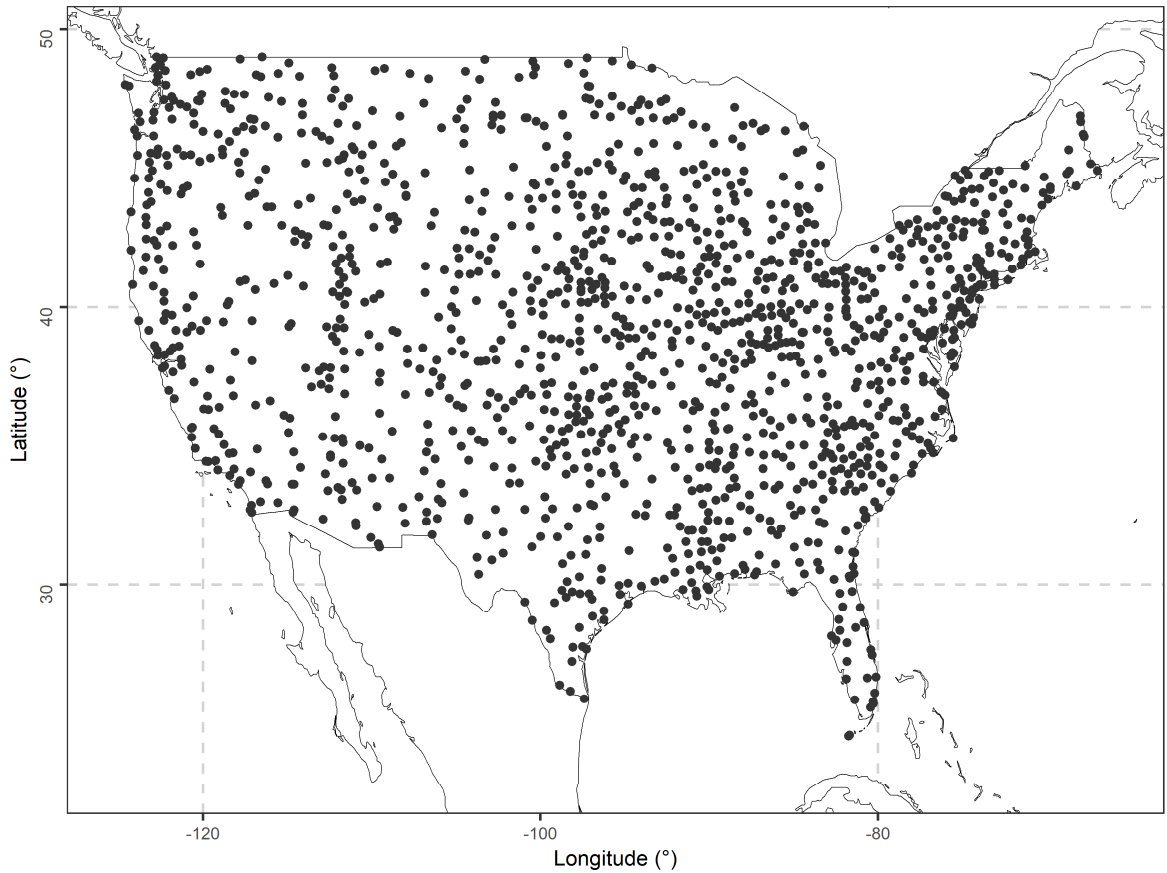

Figure 2. Map of ground-based stations that provided data for this work.

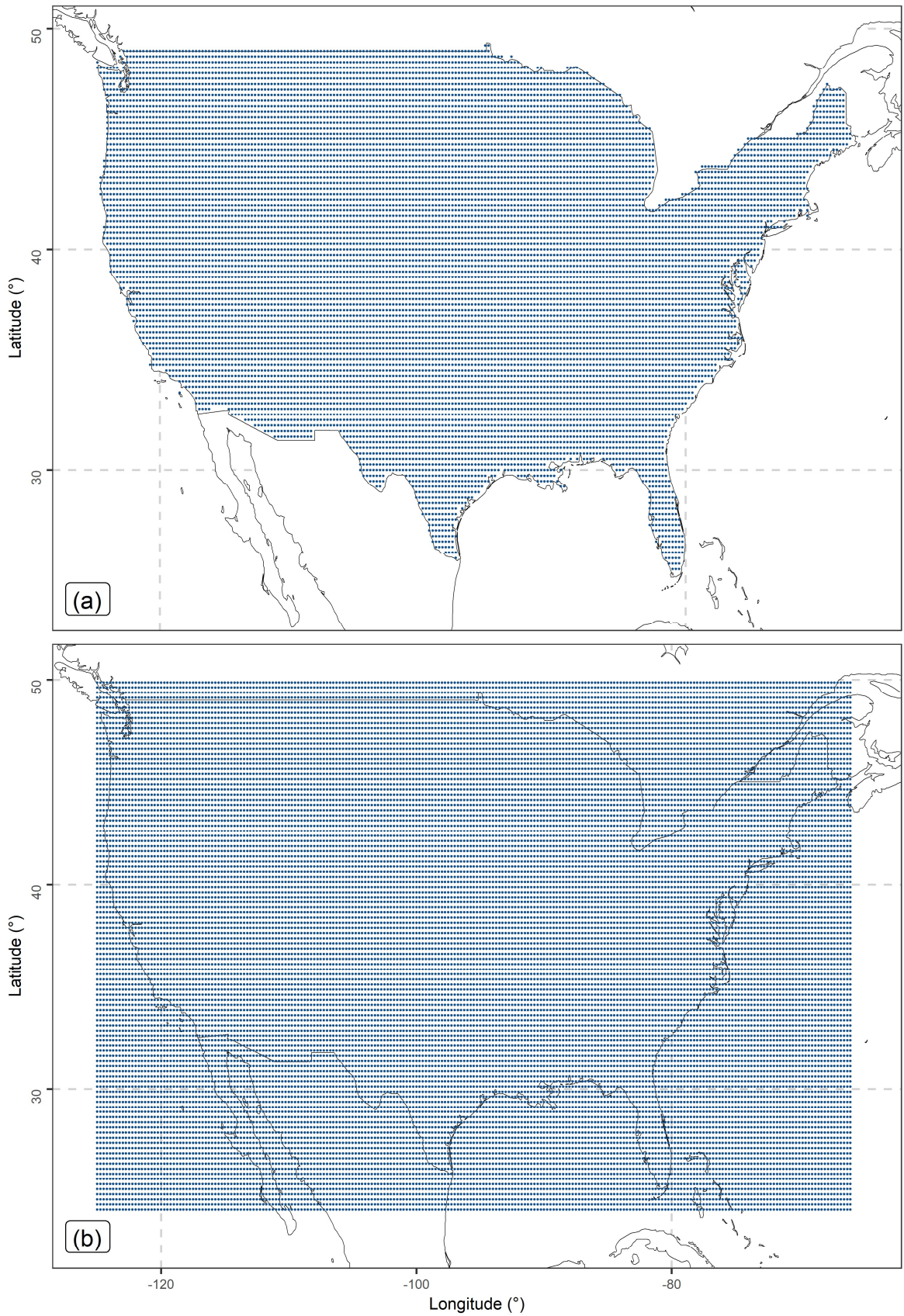

Figure 3. Maps of the (a) PERSIANN and (b) IMERG grids that provided data for this work.

More precisely, the data utilized originate from the following databases, which were also synthesized by previous works (Papacharalampous et al., 2023b, 2024a, 2024b) for benchmarking other algorithms:

o   The Global Historical Climatology Network monthly database, version 2 (GHCNm), which offers monthly precipitation data obtained through the operation of ground-based stations (Peterson & Vose, 1997).

o   The Precipitation Estimation from Remotely Sensed Information using Artificial Neural Networks (PERSIANN) database, which offers remote sensed gridded precipitation data at the daily timescale (Hsu et al., 1997; Nguyen et al., 2018; Nguyen et al., 2019).

o   The GPM Integrated Multi-satellitE Retrievals late precipitation L3 1 day 0.1° x 0.1° V06 (IMERG), which offers remote sensed gridded precipitation data at the daily timescale (Huffman et al., 2019).

o   The Amazon Web Services Terrain Tiles (AWSTT) database, which offers elevation data.

These databases were accessed, respectively, through the following sources and links:

o   The National Oceanic and Atmospheric Administration (NOAA) on 2022-09-24 through https://www.ncei.noaa.gov/pub/data/ghcn/v2.

o   The Centre for Hydrometeorology and Remote Sensing (CHRS), University of California, Irvine (UCI) on 2022-03-07 through https://chrsdata.eng.uci.edu.

o   The National Aeronautics and Space Administration (NASA) Goddard Earth Sciences (GES) Data and Information Services Center (DISC) on 2022-12-10 through https://doi.org/10.5067/GPM/IMERGDL/DAY/06.

o   The Amazon Web Services (AWS) on 2022-09-25 through https://registry.opendata.aws/terrain-tiles.

The gauge-measured precipitation data and the elevation data were used in their original forms. The same does not hold for the remote sensed data, which were daily originally and, therefore, they had to be aggregated to match the temporal resolution of the gauge-measured ones. Additionally, the IMERG data had to undergo bilinear interpolation to match the spatial resolution of the PERSIANN data.

## 3.2 Feature engineering

This study adopted the feature engineering strategy proposed in Papacharalampous et al. (2024b). This strategy relies on the distance-based weighting of the remote sensing data, thereby reducing to half, with limited loss of information, the number of satellite-based predictor variables compared to previous studies (e.g., Papacharalampous et al., 2024a).

In detail, the dataset included 91 623 samples. Each of the latter had the form $\text{sample}_k$ = {$\text{PR}_{\text{station}}$, $\dot{\text{PR}}_{1,\text{PERSIANN}}$, $\dot{\text{PR}}_{2,\text{PERSIANN}}$, $\dot{\text{PR}}_{3,\text{PERSIANN}}$, $\dot{\text{PR}}_{4,\text{PERSIANN}}$, $\dot{\text{PR}}_{1,\text{IMERG}}$, $\dot{\text{PR}}_{2,\text{IMERG}}$, $\dot{\text{PR}}_{3,\text{IMERG}}$, $\dot{\text{PR}}_{4,\text{IMERG}}$, $\text{elevation}_{\text{station}}$}, $k = 1, \ldots, 91\,623$, where:

- $\text{PR}_{\text{station}}$ is the precipitation value from a ground-based station at a specified time point, which refers to a specific pair {month, year}.

- $\dot{\text{PR}}_{i,\text{PERSIANN}}$ and $\dot{\text{PR}}_{i,\text{IMERG}}$, $i = 1, \ldots, 4$, are the distance-weighted remote sensing precipitation values of the four closest PERSIANN grid points and the four closest IMERG grid points, respectively, at the same time point. More precisely, the distance-weighted precipitation $\dot{\text{PR}}_i$ at the grid point $i = 1, \ldots, 4$ is defined as

$$\dot{\text{PR}}_i := \frac{(1/d_i^2)\text{PR}_i}{\sum_{j=1}^{4} 1/d_j^2}, i = 1, \ldots, 4, \quad (5)$$

where $\text{PR}_i$, $i = 1, \ldots, 4$, are the raw values of remote sensing precipitation at the four closest grid points and $d_i$, $i = 1, \ldots, 4$, are the distances of the station from the four closest grid points (Figure 4).

- $\text{elevation}_{\text{station}}$ is the elevation at the location of the station.

Among the values contained in each sample, $\text{PR}_{\text{station}}$ represented the dependent variable and the others represented the predictor variables.

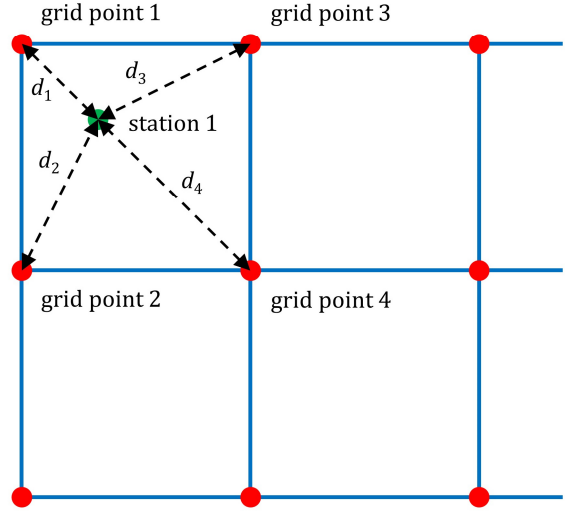

Figure 4. How the data samples for the application of the algorithms were formed. Stations and, thus, station-measured precipitation data were available for the locations shown in Figure 2, while remote sensing precipitation data were available for the PERSIANN and IMERG grid points shown in Figure 3. The predictand was station-measured precipitation. For each station, data for eight predictor variables were formed following distance-based weighting of the remote sensing precipitation values at the four closest grid points (i.e., the grid points 1–4 for station 1). Data for a ninth predictor variable were obtained by estimating the elevation of the station.

## 3.3  Algorithm testing

We randomly divided the 91 623 samples into three equally sized sets. On the first set, we trained the six base learners (see Section 2.2) to subsequently apply them to obtain predictions for the second test. From the predictive probability distributions in the second test, we extracted the predictions for the quantile levels $\tau \in \{0.0125, 0.025, 0.050, 0.075, 0.100, 0.200, 0.300, 0.400, 0.500, 0.600, 0.700, 0.800, 0.900, 0.925, 0.950, 0.975, 0.9875\}$. While we could extract the predictions for any quantile level, we selected these, as they consist a good approximation of the predictive probability distribution, which also emphasizes on the lowest and the extreme quantiles.

Then, we used these predictions as predictor variables within the frameworks of the stacked generalization algorithms (see Sections 2.2 and 2.3) to predict the true values (see the pseudo algorithm in Section 2.3). Subsequently, we trained the base learners on the union of sets 1 and 2, and used them to obtain predictions for the third set. We used these latter predictions to form the predictions of the ensemble learners for the third set, which served for testing purposes (see again the pseudo algorithm in Section 2.3). We also used them to benchmark the ensemble learners. For benchmarking purposes, we additionally applied the two other individual algorithms on the union of sets 1 and 2, and used them to obtain predictions for the test set.

### 3.4 Evaluation metrics

According to Equation (4), we computed the quantile loss in the test set for each pair {algorithm, quantile level}. Among two predictions of the $\tau$-quantile, the one with lowest loss is ranked first. Subsequently, we obtained a median quantile score for each algorithm over the test set according to the following equation:

$$\bar{L}_\tau(z, y) := \text{median}_n\{L_\tau(z_i, y_i)\}, \tag{6}$$

where $\tau$ is the quantile level of interest, $L_\tau$ is the quantile loss function defined by Equation (4), $n$ is the size of the test set, and $y_i$ and $z_i$, $i \in \{1, \dots, n\}$ are the observation and $\tau$-quantile prediction, respectively, of the $i^{\text{th}}$ sample.

$\bar{L}_\tau(z, y)$ takes values between 0 and $\infty$. The lower its value, the better the prediction, while values of $\bar{L}_\tau(z, y)$ equal to 0 correspond to optimal predictions of the $\tau$-quantile. Because the median quantile scores are not scaled, we computed quantile prediction skill scores. These scores take values between $-\infty$ and 1, they allow to understand the relative improvement of the method of interest with respect to a (simple) reference method and are obtained according to the following equation (Gneiting, 2011):

$$L_{\tau,\text{skill}} := 1 - \bar{L}_{\tau,\text{algorithm}} / \bar{L}_{\tau,\text{reference}}, \tag{7}$$

We defined the reference method as the $\tau$-quantile of the training set. The predictions are optimal when the quantile prediction skill is equal to 1 (i.e. when $\bar{L}_{\tau,\text{algorithm}} = 0$), and better (worse) than the predictions of the reference method, when the quantile prediction skill is larger (smaller) than 0. A higher quantile prediction skill score $L_{\tau,\text{skill}}$ indicates a better prediction. To ease the comparison, ranks of the algorithms based on the quantile prediction skill were also computed.

We also summed up, separately for each pair {sample, algorithm}, the quantile losses obtained at various quantile levels to obtain values of quantile scoring rules (Cervera & Muñoz, 1996; Gneiting & Raftery, 2007). Scoring rules allow comparing the full probabilistic predictions. The quantile scoring rule is proper, i.e. its expected score is minimized when the true probability distribution is issued. Proper scoring rules allow ranking of probabilistic predictions, with lower values of the scoring rule indicating better performance. Practically, they are loss functions, where the prediction argument is given in the form of a probability distribution. For the case of the quantile scoring rule, the probability distribution is given in the form of predictive quantiles at multiple levels (quantile levels of interest were listed in Section 3.3).

Furthermore, we averaged, separately for each algorithm, the sums across all the samples to take quantile scoring rule skills, using as reference method the $\tau$-quantiles of the training set, in a similar way to Equation (7),

In addition to quantile prediction skill and quantile scoring rule skill, we computed the sample coverage for each pair of {algorithm, quantile level}. Sample coverage refers to the frequency with which a prediction is less than or equal to its corresponding true value. Sample coverages closer to the nominal quantile levels indicate more reliable predictions and suggest good absolute performance (Fissler et al., 2021). However, if two algorithms have good absolute performance, we would prefer the one with the lowest quantile score. This is because a lower quantile score rewards both the reliability and sharpness of the prediction (Gneiting & Raftery, 2007). Sharpness refers to the ability to provide narrow prediction intervals, which is particularly important when prediction intervals are of interest. The concept of sharpness also applies to the prediction of quantiles.

## 4. Results

Figure 5 presents the differences in the performance of the algorithms in terms of the quantile scoring rule skill. Therefore, it allows comparisons with respect to the entire predictive probability distributions of precipitation, which are represented in this work by quantile predictions at multiple quantile levels. The algorithm with the larger quantile scoring rule skill is the stacking of GAMLSS-ZAIG-Splines, GAMLSS-ZAGA-Splines, DRF-ZAIG and DRF-ZAGA. The stacking of GAMLSS-ZAGA-Splines and DRF-ZAGA follows very closely with an (almost) identical score, and the same holds for the stacking of DRF-ZAIG and DRF-ZAGA. More generally, in the problem investigated, the stacking strategy beats both the mean and the median combiners independently of the algorithms combined.

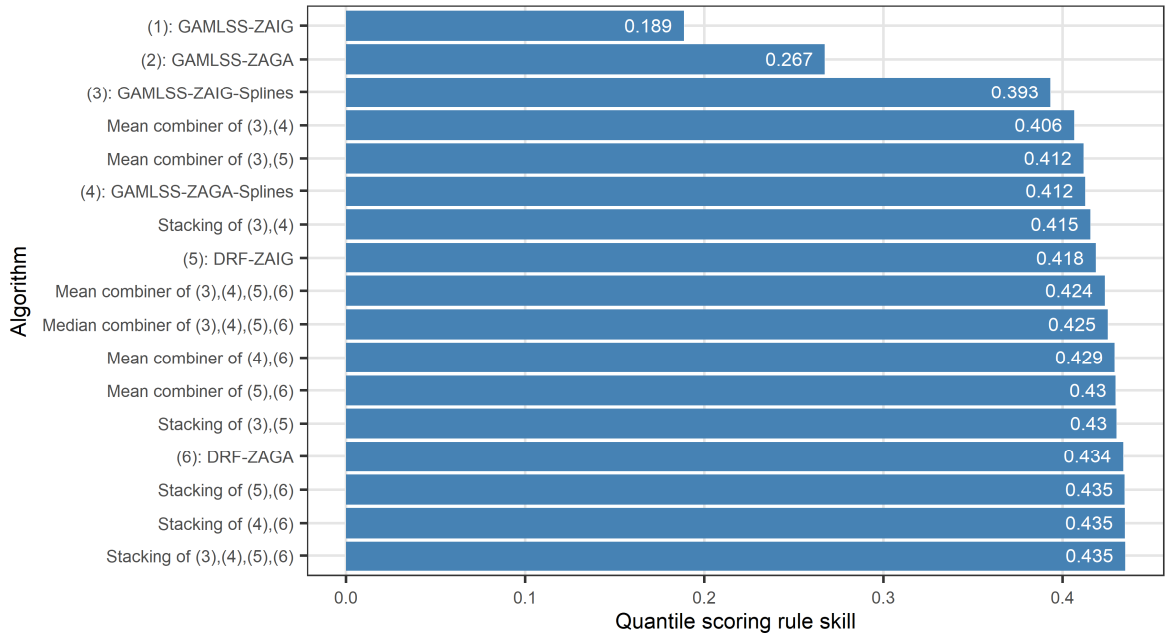

Figure 5. Quantile scoring rule skill of the algorithms. The larger the quantile scoring rule skill, the better the probabilistic predictions on average compared to the "climatology" predictions. The algorithms are listed in their order from the worst (top) to the best (bottom).

It is also worth mentioning that the ZAGA distribution offers better skill than the ZAIG one, overall, independently on whether it is used into a GAMLSS or a distributional regression algorithm. Still, the strategy of taking advantage of both these distributions through combination methods proposed in this work (e.g., through stacking the quantile regression algorithm as the combiner on the top of GAMLSS-ZAIG-Splines, GAMLSS-ZAGA-Splines, DRF-ZAIG and DRF-ZAGA algorithms) is more likely to issue the best probabilistic predictions. Also, the benchmark methods (i.e., GAMLSS-ZAIG and GAMLSS-ZAGA) perform significantly worse than the base learners (i.e., GAMLSS-ZAIG-Splines, GAMLSS-ZAGA-Splines, DRF-ZAIG and DRF-ZAGA), a fact that perhaps indicates an advantage of the non-linear algorithms over the linear ones in the context investigated. Additionally, both versions of the distributional regression forests (i.e., DRF-ZAIG and DRF-ZAGA) perform better than both the versions of the GAMLSS algorithm with splines (i.e., GAMLSS-ZAIG-Splines and GAMLSS-ZAGA-Splines).

Figure 6 presents the differences in the performance of the algorithms in terms of the quantile prediction skill and the respective ranks. Thus, it can facilitate more in depth comparisons, which take place independently at each quantile level. Overall, there is no algorithm beating its candidates at all the quantile levels and, therefore, even the best performing algorithms in terms of quantile scoring rule skill perform worse than others at a few quantile levels. For instance, at the two lowest quantile levels (i.e., 0.0125 and 0.025), the best performance is exhibited by DRF-ZAIG (ranked 8th in terms of quantile scoring rule skill) and in the three highest (most extreme) quantiles (i.e., 0.950, 0.975 and 0.9875) the best performance is exhibited by the stacking of GAMLSS-ZAIG-Splines and GAMLSS-ZAGA-Splines (ranked 7th in terms of quantile scoring rule skill). Furthermore, while individual methods exhibit substantial fluctuations in their rankings across quantile levels, stacking methods and mean combiners generally demonstrate more stable performance. For example, DRF-ZAGA performs comparably to stacking methods with respect to the quantile scoring rule, yet it displays significant performance fluctuations across quantile levels, with ranks ranging from 2 to 15. In contrast, stacking of GAMLSS-ZAIG-Splines, GAMLSS-ZAGA-Splines, DRF-ZAIG, and DRF-ZAGA ranks between 1 and 10.5.

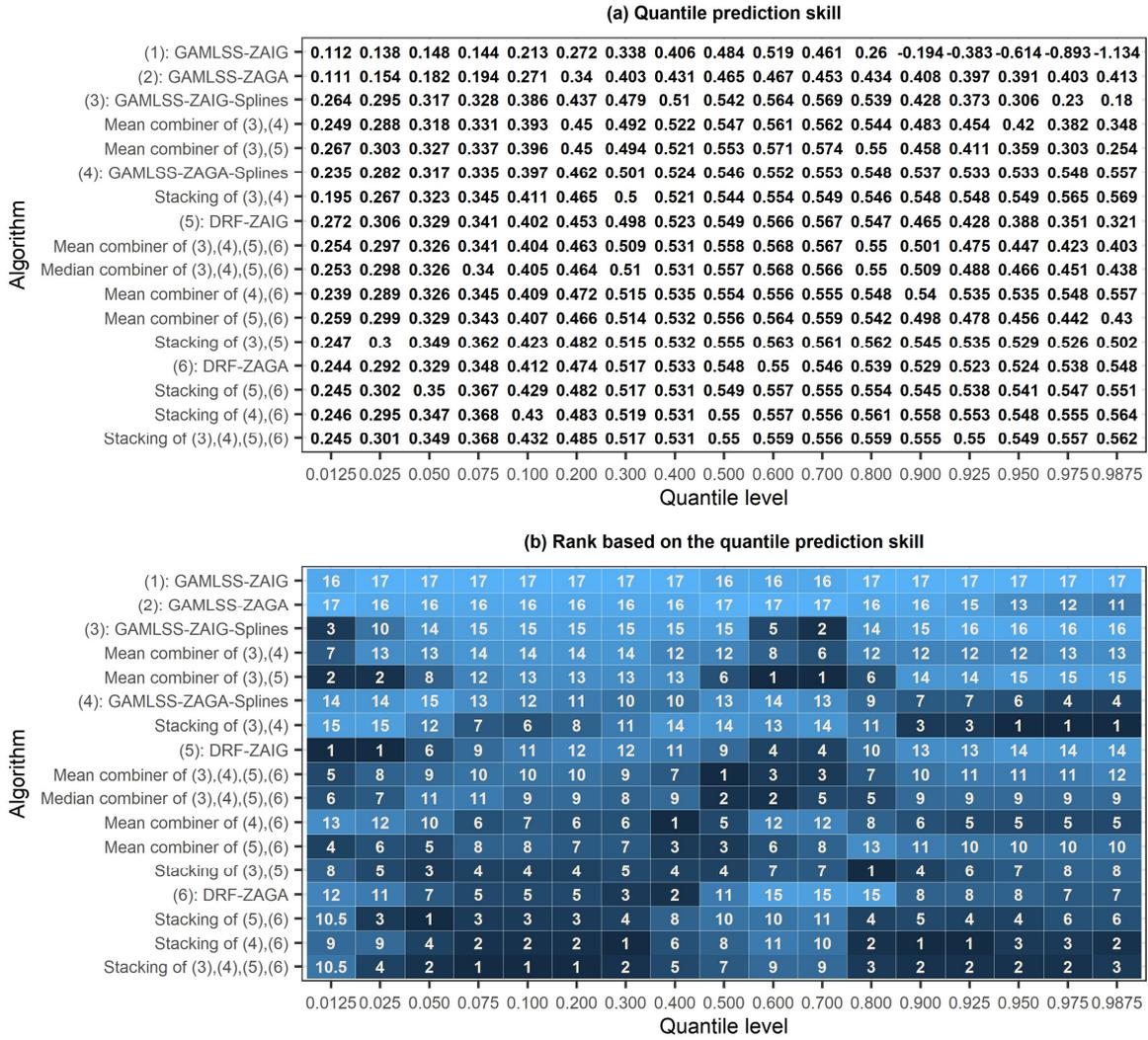

Figure 6. (a) Quantile prediction skill and (b) the respective ranks of the algorithms from the best (1st and darkest blue coloured) to the worst (17th and lightest blue coloured) at each quantile level. The larger the quantile prediction skill, the better the predictions compared to the "climatology" predictions. The algorithms are listed in their order from the worst (top) to the best (bottom) based on the quantile scoring rule skill (Figure 5).

Especially in the two highest quantiles (i.e., 0.975 and 0.9875), the differences in performance between the stacking of GAMLSS-ZAIG-Splines and GAMLSS-ZAGA-Splines and the two best-performing algorithms in terms of quantile scoring rule skill are also considerable. Furthermore, combinations based on the mean combiner appear in the first position at four central quantile levels (i.e., 0.400, 0.500, 0.600, 0.700), beating the stacked generalization methods relying on the same base learners. Lastly, the sample coverage offered by the algorithms at the various quantile levels is in a relatively good agreement with the nominal values (Figure 7), a fact indicating the reliability of the algorithms.

| Algorithm | 0.0125 | 0.025 | 0.050 | 0.075 | 0.100 | 0.200 | 0.300 | 0.400 | 0.500 | 0.600 | 0.700 | 0.800 | 0.900 | 0.925 | 0.950 | 0.975 | 0.9875 |
|---|---|---|---|---|---|---|---|---|---|---|---|---|---|---|---|---|---|
| (1): GAMLSS-ZAIG | 0.035 | 0.039 | 0.048 | 0.055 | 0.063 | 0.101 | 0.156 | 0.232 | 0.343 | 0.501 | 0.695 | 0.871 | 0.968 | 0.979 | 0.988 | 0.994 | 0.996 |
| (2): GAMLSS-ZAGA | 0.034 | 0.04 | 0.054 | 0.07 | 0.086 | 0.167 | 0.263 | 0.364 | 0.476 | 0.589 | 0.705 | 0.819 | 0.925 | 0.948 | 0.968 | 0.983 | 0.992 |
| (3): GAMLSS-ZAIG-Splines | 0.047 | 0.055 | 0.068 | 0.082 | 0.095 | 0.153 | 0.22 | 0.306 | 0.407 | 0.528 | 0.669 | 0.816 | 0.94 | 0.961 | 0.977 | 0.99 | 0.996 |
| Mean combiner of (3),(4) | 0.043 | 0.052 | 0.068 | 0.084 | 0.1 | 0.17 | 0.253 | 0.351 | 0.456 | 0.571 | 0.694 | 0.818 | 0.928 | 0.952 | 0.971 | 0.986 | 0.993 |
| Mean combiner of (3),(5) | 0.044 | 0.053 | 0.068 | 0.082 | 0.096 | 0.154 | 0.224 | 0.308 | 0.411 | 0.532 | 0.669 | 0.814 | 0.942 | 0.963 | 0.98 | 0.992 | 0.997 |
| (4): GAMLSS-ZAGA-Splines | 0.04 | 0.049 | 0.067 | 0.085 | 0.105 | 0.189 | 0.287 | 0.395 | 0.503 | 0.613 | 0.716 | 0.819 | 0.911 | 0.935 | 0.956 | 0.976 | 0.986 |
| Stacking of (3),(4) | 0.035 | 0.045 | 0.068 | 0.092 | 0.118 | 0.217 | 0.318 | 0.409 | 0.502 | 0.602 | 0.703 | 0.801 | 0.9 | 0.924 | 0.95 | 0.974 | 0.986 |
| (5): DRF-ZAIG | 0.045 | 0.055 | 0.071 | 0.085 | 0.1 | 0.159 | 0.228 | 0.311 | 0.412 | 0.527 | 0.658 | 0.803 | 0.933 | 0.957 | 0.977 | 0.992 | 0.996 |
| Mean combiner of (3),(4),(5),(6) | 0.041 | 0.05 | 0.067 | 0.084 | 0.102 | 0.171 | 0.256 | 0.352 | 0.461 | 0.575 | 0.697 | 0.82 | 0.932 | 0.953 | 0.972 | 0.988 | 0.995 |
| Median combiner of (3),(4),(5),(6) | 0.041 | 0.05 | 0.067 | 0.084 | 0.101 | 0.172 | 0.256 | 0.351 | 0.458 | 0.571 | 0.695 | 0.818 | 0.93 | 0.951 | 0.97 | 0.987 | 0.994 |
| Mean combiner of (4),(6) | 0.038 | 0.048 | 0.066 | 0.086 | 0.106 | 0.192 | 0.289 | 0.397 | 0.509 | 0.615 | 0.719 | 0.823 | 0.917 | 0.939 | 0.959 | 0.978 | 0.988 |
| Mean combiner of (5),(6) | 0.041 | 0.051 | 0.07 | 0.086 | 0.103 | 0.176 | 0.261 | 0.355 | 0.461 | 0.571 | 0.686 | 0.811 | 0.925 | 0.948 | 0.969 | 0.987 | 0.994 |
| Stacking of (3),(5) | 0.015 | 0.03 | 0.077 | 0.098 | 0.118 | 0.212 | 0.306 | 0.407 | 0.507 | 0.601 | 0.699 | 0.798 | 0.901 | 0.927 | 0.95 | 0.974 | 0.988 |
| (6): DRF-ZAGA | 0.038 | 0.048 | 0.069 | 0.088 | 0.109 | 0.197 | 0.295 | 0.399 | 0.505 | 0.609 | 0.713 | 0.818 | 0.913 | 0.936 | 0.957 | 0.977 | 0.987 |
| Stacking of (5),(6) | 0.039 | 0.052 | 0.078 | 0.099 | 0.123 | 0.213 | 0.313 | 0.409 | 0.503 | 0.601 | 0.696 | 0.796 | 0.901 | 0.927 | 0.95 | 0.975 | 0.987 |
| Stacking of (4),(6) | 0.038 | 0.049 | 0.075 | 0.1 | 0.122 | 0.216 | 0.316 | 0.404 | 0.503 | 0.6 | 0.696 | 0.796 | 0.9 | 0.927 | 0.952 | 0.976 | 0.987 |
| Stacking of (3),(4),(5),(6) | 0.038 | 0.051 | 0.077 | 0.1 | 0.123 | 0.216 | 0.318 | 0.407 | 0.504 | 0.6 | 0.697 | 0.797 | 0.9 | 0.927 | 0.951 | 0.976 | 0.987 |

Quantile level

Figure 7. Sample coverage offered by the algorithms at each quantile level. The closest the sample coverage to the nominal value (equal to $\tau$ for the quantile level $\tau$), the larger the reliability of the algorithm. The algorithms are listed in their order from the worst (top) to the best (bottom) based on the quantile scoring rule skill (Figure 5).

## 5.  Discussion

This work makes two key methodological contributions. First, it contributes to the field of (precipitation) dataset creation through data merging by proposing for the first time for the respective tasks the concept of distributional regression and several algorithms that benefit from this concept, together with zero-adjusted distributions that are befitting for modelling intermittency. Second, it contributes to the broader fields of machine learning and statistical learning by introducing and extensively comparing several ensemble learning methods that are distributional regression based and could be used in a variety of modelling contexts.

Such modelling contexts could include those of probabilistic forecasting in diverse fields (see, e.g., in Barraza et al., 2004; Chen et al., 2020; David et al., 2016; Pinson et al., 2012; Quilty et al., 2019; Taylor & Taylor, 2023; Tyralis & Papacharalampous, 2021; Wan et al., 2013), post-processing of physics-based model predictions in meteorology (see, e.g., in Grönquist et al., 2021; Medina & Tian, 2020; Phipps et al., 2022) and hydrology (see, e.g., in Bogner et al., 2016; Montanari & Koutsoyiannis, 2012; Papacharalampous & Tyralis, 2022; Tyralis & Papacharalampous, 2023a), as well as numerous spatial prediction problems beyond the focus of this work (see, e.g., in Fendrich et al., 2024; Schmidinger & Heuvelink, 2023).

In such technical problems, the new methods could be compared experimentally with the current state-of-the-art ones, especially with those that are designed for predicting extreme quantiles and/or modelling intermittency or those that are parametric or semi-parametric and, thus, can take advantage of knowledge at hand about which distribution describes the predictand better (Tyralis & Papacharalampous, 2024). Comparisons with non-parametric methods could also be useful, given that they will be made in a way that does not diminish the different advantages of the different method families. Importantly, different distributions could also be useful components of the new methods in different modelling contexts, as well as in the problem of focus here but at different time scales.

Non-parametric algorithms, such as linear quantile regression (Koenker, 2005; Koenker & Bassett Jr, 1978), boosting quantile regression and their variants (Friedman, 2001; Ke et al., 2017; Mayr et al., 2014), quantile regression forests and their variants (Athey et al., 2019; Meinshausen & Ridgeway, 2006) and quantile regression neural networks (Cannon, 2011; Taylor, 2000), might outperform distributional regression in terms of prediction performance, while they have already been used in precipitation spatial prediction (Papacharalampous et al., 2024a, 2024b). This is because the form of the conditional probability distribution is often misspecified in real-world situations. However, distributional regression algorithms can explicitly model intermittency, which is an important property of precipitation. This allows for more realistic modelling compared to non-parametric approaches. Additionally, modelling the full conditional probability distribution enables effortless extrapolation to high extreme quantiles. In contrast, extrapolating to high quantiles with non-parametric models requires imposing an extreme value distribution (Tyralis et al., 2023a; Wang et al., 2012; Wang & Li, 2013).

The stacking strategy proposed in this work outperforms both individual algorithms and simple averaging. However, it has some limitations. In particular, the weights for each algorithm need to be estimated at every quantile level of interest. This can become problematic at higher quantiles, especially with small datasets. In such cases, the quantile loss function may not provide adequate training for predicting data beyond the observed range (Gandy et al., 2022). For these scenarios, simple averaging might be a preferable solution, as it has been shown to maintain high performance (Petropoulos & Svetunkov, 2020; Smith & Wallis, 2009). Alternatively, methods that combine full probability distributions can be used to address issues arising from quantile-based combinations (Gneiting & Ranjan, 2013).

## 6. Summary and conclusions

This work focused on the important spatial prediction problem of estimating predictive uncertainty while merging satellite and gauge precipitation datasets. It introduced the concept of distributional regression for solving this problem. Additionally, it identified two distributions, namely the zero adjusted inverse Gaussian (ZAIG) and zero adjusted gamma (ZAGA), as useful components of distributional regression methods when dealing with the task of focus at the monthly temporal scale, and formulated ensemble learning methods that could also be used for solving other spatial prediction and, more generally, prediction problems.

The new ensemble learning methods benefit from spline-based generalized additive models for location scale and shape (GAMLSS), distributional regression forests, quantile regression and three ensemble learning strategies, specifically the stacking one and the mean and median combiners. They also benefit from the aforementioned distributions. With respect to quantile regression-based algorithms, which are the current state-of-the-art in the area of interest, the methods of this work, with them including the new ones and several benchmarks, are expected to perform better both at modelling intermittency and at extrapolating at high quantiles.

A detailed comparison of these methods was conducted by taking advantage of a big multisource dataset, as well as the theory of scoring functions and scoring rules. Three methods were identified as the best in terms of the quantile scoring rule skill. These are the following: (a) stacking of spline-based GAMLSS with ZAIG, spline-based GAMLSS with ZAGA, distributional regression forests with ZAIG and distributional regression forests with ZAGA; (b) stacking of spline-based GAMLSS with ZAGA and distributional regression forests with ZAGA; and (c) stacking of distributional regression forests with ZAIG and distributional regression forests with ZAGA.

Still, the results also suggest the usefulness of other methods among the new or even the benchmark ones. Indeed, the above three methods were beaten by others at several quantile levels. For instance, at the lowest quantile levels, distributional regression forests with ZAIG were ranked first in terms of the quantile prediction skill. Other examples refer to the most extreme and the central quantiles, at which the best performance was exhibited, respectively, by the stacking of spline-based GAMLSS with ZAIG and spline-based GAMLSS with ZAGA, and by combinations based on the mean combiner. It is important to note, however, that the stacking methods, and to a lesser extent the mean combiners, exhibit less variability in their performance across quantile levels compared to individual methods that occasionally underperform significantly. Overall, a task-specific utilization of multiple algorithms seems to be needed to obtain the best possible predictive performance.

**Declaration of competing interest:** There is no conflict of interest.

**Declaration of generative AI in scientific writing:** During the preparation of this work, the authors used Gemini to improve readability and language. After using this tool, the authors reviewed and edited the content as needed and take full responsibility for the content of the publication.

**Funding:** This work was conducted in the context of the research project BETTER RAIN (BEnefiTTing from machine lEarning algoRithms and concepts for correcting satellite RAINfall products). This research project was supported by the Hellenic Foundation for Research and Innovation (H.F.R.I.) under the "3rd Call for H.F.R.I. Research Projects to support Post-Doctoral Researchers" (Project Number: 7368).

**Acknowledgements:** We thank the Associate Editor and the Reviewers for their helpful comments on the first version of the manuscript.

**Appendix A    Statistical software**

We programmed the ensemble learning algorithms and conducted the experiments of this study in the `R` programming language (R Core Team 2024). The individual distributional and quantile regression algorithms were implemented through the `disttree` (Schlosser et al. 2021), `gamlss` (Rigby and Stasinopoulos 2005, Stasinopoulos and Rigby 2024) and `quantreg` (Koenker 2023) contributed `R` packages. The computation of the scoring functions was made through the `scoringfunctions` (Tyralis and Papacharalampous 2023b, 2024) contributed `R` package. The data processing, report production and figure preparation were made through the `caret` (Kuhn 2023), `data.table` (Barrett et al. 2023), `devtools` (Wickham et al. 2022), `elevatr` (Hollister 2023), `knitr` (Xie 2014, 2015, 2023), `ncdf4` (Pierce 2023), `rgdal` (Bivand et al. 2023), `rmarkdown` (Allaire et al. 2023, Xie et al. 2018, 2020), `sf` (Pebesma 2018, 2023), `spdep` (Bivand 2023, Bivand and Wong 2018, Bivand et al. 2013) and `tidyverse` (Wickham et al. 2019, Wickham 2023) contributed `R` packages.